\documentclass[preprint,12pt]{elsarticle}

\usepackage{amssymb}
\usepackage{graphicx}
\usepackage{mathtools}
\usepackage{makecell}
\usepackage{color}
\usepackage{graphicx}
\usepackage{subfigure}
\usepackage{color}
\usepackage[colorlinks, linkcolor=red,  anchorcolor=red, citecolor=blue]{hyperref}
\usepackage{textcomp,booktabs}
\usepackage{amssymb}
\usepackage{pifont}
 \usepackage{amssymb}
\usepackage{colortbl}
\definecolor{mygray}{gray}{.9}
\definecolor{mypink}{rgb}{.99,.91,.95}
\definecolor{mycyan}{cmyk}{.3,0,0,0}

\begin{document}

\begin{frontmatter}

\title{STADB: A Self-Thresholding Attention Guided ADB Network for Person Re-identification}

\author{Bo Jiang, Sheng Wang, Xiao Wang*, Aihua Zheng}

\address{Bo Jiang and Sheng Wang are from Anhui Provincial Key Laboratory of Multimodal Cognitive Computation, School of Computer Science and Technology of Anhui University, Hefei,230601, China. 
Aihua Zheng is from School of Artificial Intelligence, Anhui University, Hefei, 230601, China. 
Xiao Wang is from Peng Cheng Laboratory, Shenzhen, China. * Corresponding author: Xiao Wang. }

\begin{abstract}
Recently, Batch DropBlock network (BDB) has demonstrated its effectiveness on person image representation and re-identification task via feature erasing. However, BDB drops the features \textbf{randomly} which may lead to sub-optimal results. In this paper, we propose a novel Self-Thresholding attention guided Adaptive DropBlock network (STADB) for person re-ID which can \textbf{adaptively} erase the most discriminative regions. Specifically, STADB first obtains an attention map by channel-wise pooling and returns a drop mask by thresholding the attention map. Then, the input features and self-thresholding attention guided drop mask are multiplied to generate the dropped feature maps. In addition, STADB utilizes the spatial and channel attention to learn a better feature map and iteratively trains the feature dropping module for person re-ID. Experiments on several benchmark datasets demonstrate that the proposed STADB outperforms many other related methods for person re-ID. The source code of this paper is released at: \textcolor{red}{\url{https://github.com/wangxiao5791509/STADB_ReID}}. 
\end{abstract}


\end{frontmatter}


\section{Introduction}
\label{intro}

Given a probe image, the goal of person re-identification (re-ID) is to search the pedestrian image of the same identity from a gallery set. It has been widely used in many applications, such as video surveillance and self-driving. In recent years, many deep learning based person re-ID approaches \cite{dai2019batch, Zhong_2018_CVPR, Deng_2018_CVPR, song2018mask, liu2017end, li2018harmonious, xu2018attention, li2018diversity, bai2020deep} are proposed and achieve great improvements over traditional approaches. However, the re-ID performance in some challenging scenarios is still unsatisfied due to the influence of clutter background, illumination, motion blur, low resolution, and occlusion.


To address the above issues, many researchers resort to attention mechanisms for person re-ID \cite{liu2017end, li2018harmonious, song2018mask, xu2018attention, li2018diversity}. Attention mechanisms usually pursue to exploiting the most discriminative feature learning, which have been successfully used in many other computer vision tasks. However, how to learn robust fine-grained local features for person re-ID is still a challenging issue. To overcome this issue, some works also introduce some other additional information, such as attribute recognition \cite{wang2019pedestrian}, pose estimation \cite{saquib2018pose} and part detection \cite{sun2019perceive} to improve person re-ID performance.
Recent works also demonstrate that hard sample mining/generation strategies usually perform beneficially for robust feature learning \cite{wang2017fast, wang2018learning, Wang_2018_CVPR, Zhong_2018_CVPR, Deng_2018_CVPR, Simo2014Fracking, Xiaolong2015Unsupervised, Loshchilov2015Online, dai2019batch, choe2019attention, wang2019improved}. Among them, Batch DropBlock Network (BDB)~\cite{dai2019batch} is a recent feature learning approach that can jointly utilize the global and local feature representations for person re-ID. Specifically, it introduces a feature dropping module to randomly erase the most discriminative features and thus focuses more on non-discriminative features. However, one main limitation of BDB~\cite{dai2019batch} is that it \textbf{randomly} drops the features to generate hard samples for training the network which may be sub-optimal. Recent works \cite{wang2017fast, Wang_2018_CVPR, choe2019attention} reveal that carefully designed feature dropping module can achieve better performance. This inspires us to rethink how to drop specific regions of extracted feature maps to obtain better fine-grained local features.

Inspired by recent works \cite{choe2019attention, dai2019batch}, this paper develops a Self-Thresholding attention guided Adaptive DropBlock Network (STADB) for person re-ID, as shown in Figure \ref{fig_2}.
The key aspect of STADB is to adaptively erase the most discriminative features according to the \textbf{estimated attention map} \cite{choe2019attention} rather than \textbf{randomly} dropping in BDB~\cite{dai2019batch} for re-ID problem. More specifically, STADB mainly contains three sub-networks, i.e., global branch, attention branch, and local feature drop network.
First, we employ a global branch to extract the global feature representation for the input pedestrian image.
Second, we use a local feature dropping module to adaptively erase the most discriminative parts and make our neural network be more sensitive to the non-discriminatory features. We erase the discriminative features according to the estimated self-thresholding attention guided regions.
Finally, we introduce the spatial and channel attention for more discriminative feature representation learning for person re-ID.
This attention branch and local feature dropping network can be randomly selected and optimized along with a global branch together in the training phase. Previous works generally either utilize feature dropping module or attention module for robust feature learning, while our method makes full use of the advantages of both of them simultaneously.

\begin{figure*}[!t]
	\begin{center}
		\includegraphics[width=0.75\textwidth]{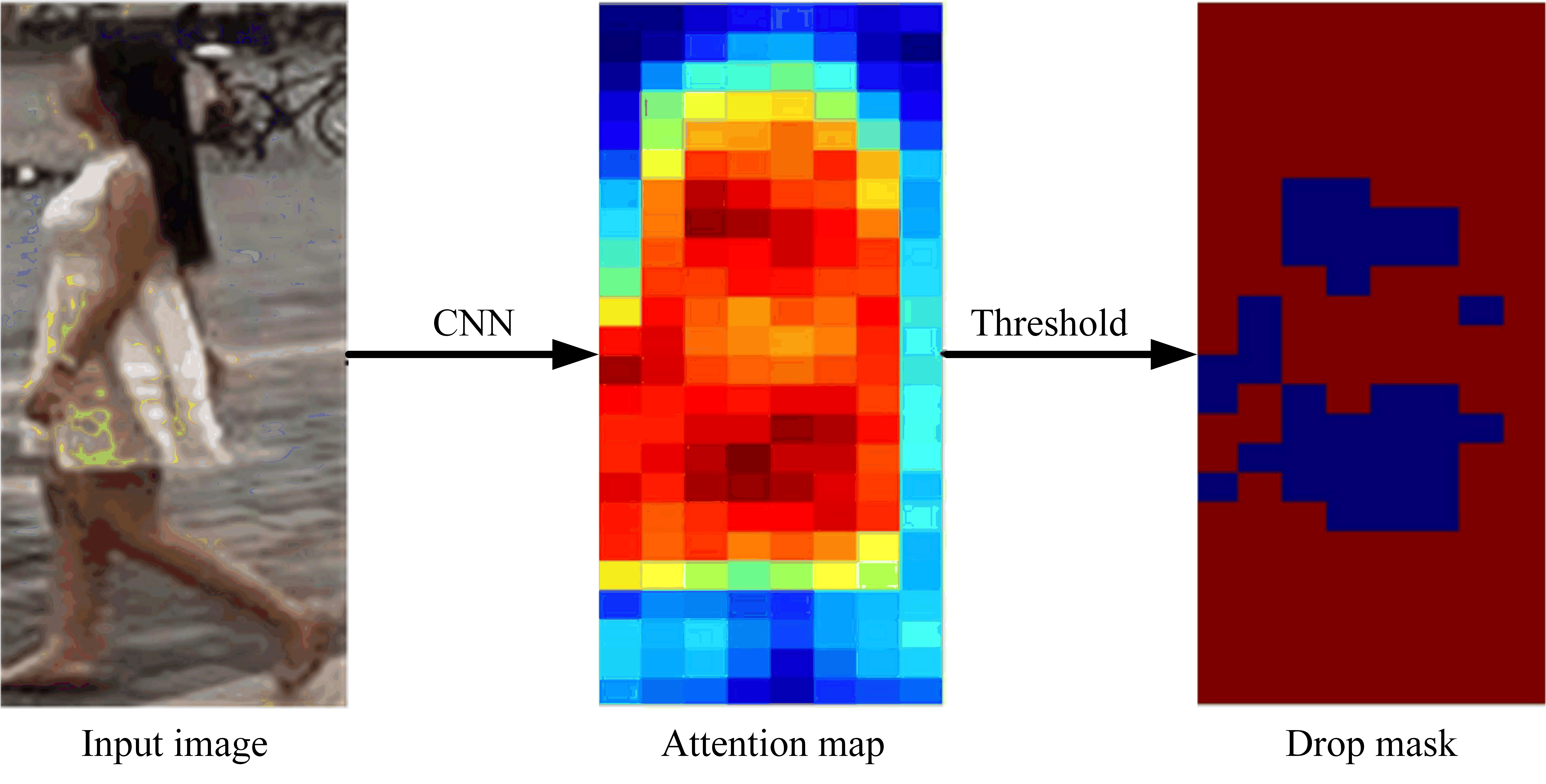}
	\end{center}
	\caption{The process of the drop mask generation. In particular, red regions in the attention map denotes more discriminative features compared with the blue regions. The elements in the drop mask are 0 and 1, where 0 on the blue regions corresponds the discriminative area in the attention and the rest area is represented by 1.}
	\label{fig_3}
\end{figure*}

The main contributions of this paper can be summarized as follows:
\begin{itemize}
	\item We propose to employ Self-Thresholding attention guided Adaptive feature Dropping Module (STADB) for the person image representation and identification tasks.
	
	\item We jointly utilize the adaptive feature dropping module and attention scheme which can attain better feature representation for person re-ID.
	
	\item Extensive experiments on multiple person re-ID benchmark datasets validate the effectiveness of the proposed STADB network.
\end{itemize}


\section{Related Work} \label{relatedWorks}

\textbf{Person Re-identification: }
 Recent person re-ID approaches generally utilize CNN to automatically learn the deep features from massive training datasets~\cite{chang2018Multi, shen2018deep, sun2017svdnet, Zhao2017Deeply,zhao2020similarity, luo2019alignedreid++,avola2019master,ren2019deep,hao2020modality, shu2021semantic, shu2021large}. Image dicing and semantic estimation technique are also popular strategies to extract local features \cite{Varior2016A,Zhao2017Spindle}. 
In addition, many works introduce some attention modules into person re-ID networks. 
For example, Liu \emph{et al.} \cite{liu2017end} demonstrate that multiple local areas with more distinguishable information could further improve the overall performance. Li \emph{et al.} \cite{li2018harmonious} propose to jointly learn the hard and soft attention for person re-ID. A continuous attention model guided by a binary mask is introduced in work \cite{song2018mask} which firstly uses the binary segmentation masks to construct the synthetic RGB-Mask pairs and then employs a mask-guided contrastive attention model to learn features separately from the body and background regions. Xu \emph{et al.} \cite{xu2018attention} introduces the pose-guided part attention (PPA) and attention-aware feature composition (AFC) for person re-ID in which PPA is used to mask out undesirable background features in person feature maps and can also handle the part occlusion issue. Li \emph{et al.}~\cite{li2018diversity} propose to use the spatial attention to handle the issue of alignment between frames. 
{Wu et al. \cite{wu2018and} propose a deep multiplicative integration gating function for re-ID. They also explore a deep attention-based spatially recursive model to attend to object parts and encode them into spatially expressive representations in \cite{wu2018deep}.  Some other works also explore the attention and discriminative feature learning  models  for person re-ID \cite{yang2019attention, chen2019mixed, chen2019self, zhou2019discriminative}. }
Although these works achieve better results, however, they all attempt to mine the most discriminative features and thus ignore the fine-grained local features, which are important cues in some challenging scenarios. In this paper, we jointly utilize the feature dropping module and attention model, which can obtain better local feature representations for person re-ID.

\textbf{Hard Example Generation: }
Some researchers attempt to design hard example mining/generation techniques \cite{choe2019attention, wang2017fast, wang2018learning, Wang_2018_CVPR, Zhong_2018_CVPR, Deng_2018_CVPR, Simo2014Fracking, Xiaolong2015Unsupervised, Loshchilov2015Online, dai2019batch, choe2019attention, wang2019improved} for person re-ID and other related computer vision tasks.
Specifically, in work \cite{choe2019attention}, the authors propose a self-attention mechanism (Attention-based Dropout Layer (ADL)) to process the feature maps of person images for re-ID. Wang \emph{et al.} \cite{wang2019improved} propose to utilize person attributes to mine hard mini-batch samples for the training of their network. Wang \emph{et al.} \cite{wang2018learning} propose to combine global features with multi-granularity local features and to characterize the integrity of input image. Dai \emph{et al.} \cite{dai2019batch} propose a Batch Dropblock Network (BDB) to learn some attentive local features for re-ID. Although the BDB network can obtain better performance for person re-ID and some other related retrieval tasks, however, the design of this module may still not be optimal. In this paper, we propose to drop the features guided by a self-thresholding attention module and design a novel STADB for person image representation and identification.

\section{The Proposed Approach}

In this section, we first give an overview of our proposed person re-ID model. Then, we provide the details of each component of our model. Finally, we present the details of the proposed model in training and testing phase.

\subsection{Overview} \label{Overview}

\begin{figure*}[!t]
	\begin{center}
		\includegraphics[width=0.95\textwidth]{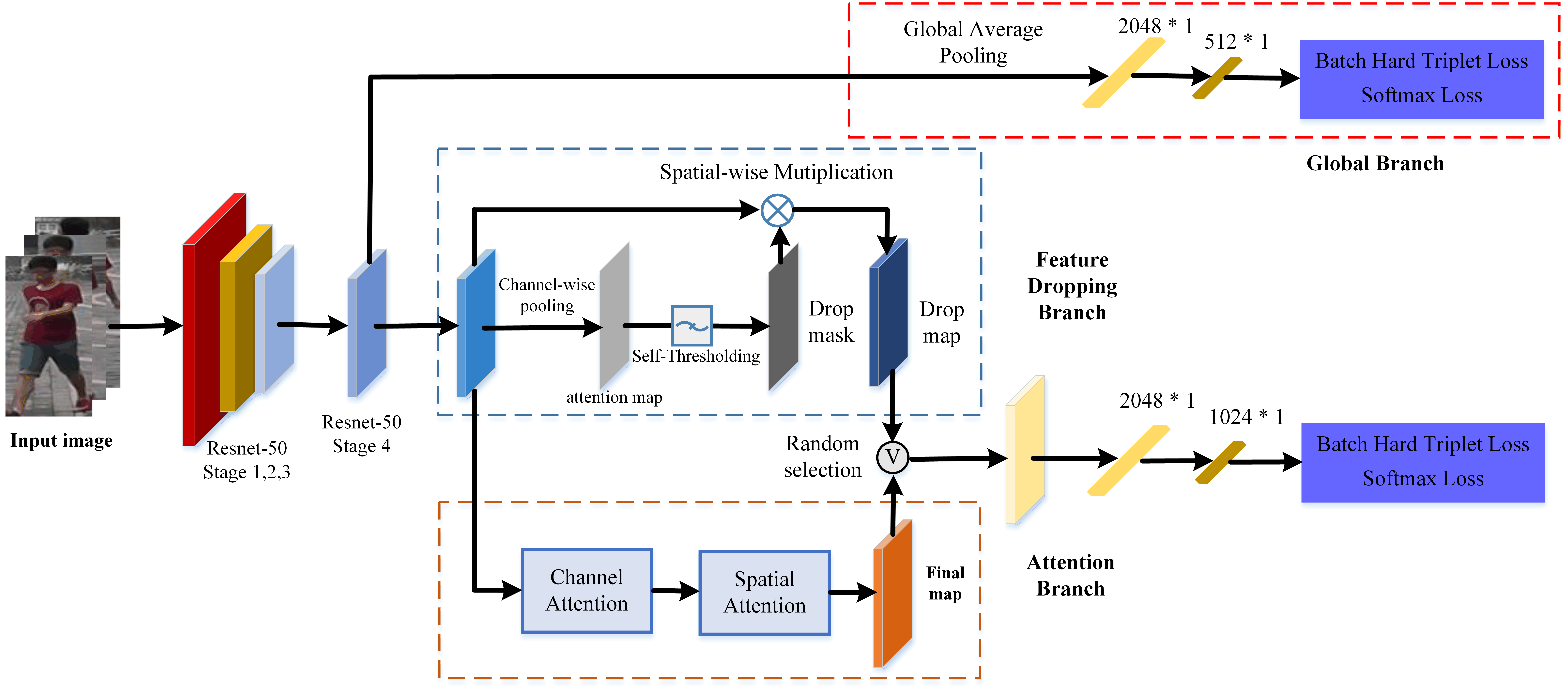}
	\end{center}
	\caption{The procedure of our proposed person re-identification network with adaptive dropblock module.}
	\label{fig_2}
\end{figure*}

As shown in Figure \ref{fig_2}, the proposed network mainly contains three modules, i.e., global branch, attention branch, and local feature drop module. The global branch is used to encode the global feature representation of the given pedestrian image. To capture the local detailed information of pedestrian images, we introduce the local feature drop network to adaptively erase the most discriminative parts by employing a self-thresholding attention scheme. In addition, we introduce the widely used spatial and channel attention modules to further improve the discriminative ability of learned feature. This attention branch and local feature drop network can be randomly selected and optimized along with the global branch. More details about these modules are described below.

\subsection{Network Architecture} \label{ourNetwork}

\subsubsection{\textbf{Global Branch}}

For person re-ID task, CNN is usually adopted for global feature extraction. As shown in Figure \ref{fig_2}, we utilize ResNet-50 \cite{he2016deepResidual} as our backbone network by following \cite{dai2019batch}. Given the feature map predicted by the backbone network, we first use a GAP (Global Average Pooling) layer to transform the feature map into a vector, followed by two fully connected layers (FC) to encode the feature vector into the fixed dimension. The number of neurons of the two FC layers is set as 2048 and 512, respectively.

\subsubsection{\textbf{Adaptive Dropping Branch} }

The motivation of this module is that 
in some challenging cases, the most discriminative features may not be the target person due to the influence of similar targets or occlusion.
Inspired by the feature dropping module proposed in previous work BDB \cite{dai2019batch} and ADL \cite{choe2019attention}, we utilize such mechanism to boost the robustness of the person re-ID model. 
Specifically speaking, as shown in Figure \ref{fig_2}, this module takes the feature maps extracted from the backbone network as the input. Then, we employ a channel-wise pooling operation on this feature map to obtain an attention map. In this way, we obtain a corresponding dropping mask via a threshold selection operation. The drop mask is used to mask the input feature map to generate the dropped feature map. The obtained feature map contains non-discriminative features that can make our neural network be more sensitive to these features, as discussed in work~\cite{choe2019attention}. The threshold is defined as follows. Assuming $x$ is the maximum pixel value in the feature map of attention map, we set the threshold as $y = \alpha  * x$ to attain a drop mask, where $\alpha$ is a hyper-parameter, * denotes the multiply operation between two real values. More concretely, the value of attention map which is greater than $y$ will be set to 0, otherwise, we set it as 1. After we obtain the drop mask, we multiply it with input feature map to obtain the drop map. Such a drop map does not contain discriminative features of the target object which encourages the neural networks to pay more attention to the non-discriminative features.

\subsubsection{\textbf{Attention Module} }

In addition to above feature dropping module, which aims to mine the non-discriminative features, we further introduce the attention estimation to learn the most discriminative feature map, 
as utilized in many previous works~\cite{liu2017end, li2018harmonious, song2018mask, xu2018attention, li2018diversity}. We introduce spatial and channel attention~\cite{woo2018cbam} to highlight the most discriminative features for person re-ID task. Therefore, we can attain more robust feature representation by designing random selection of adaptive feature dropping module and attention module when training our neural network.

\emph{Channel Attention}:
First, we take the input feature map $F\subseteq{R^{C \times H \times W}}$ as input ($C, H$ and $W$ denote channel number, height, and width of feature map respectively) and use the feature correlation between channels to generate channel attention features. The channel information of the feature map can be aggregated with global max pooling and global average pooling operations based on width and height, respectively. Therefore, we can obtain two-channel descriptions $F_{max}$ and $F_{avg} \subseteq {R^{1 \times 1 \times C}}$ and feed them into a shared network to generate two kinds of feature descriptors $M_{max}$ and $M_{avg} \in {R^{C \times 1 \times 1}}$. The shared network consists of a multi-layer perceptron (MLP) and one hidden layer. To reduce the parameter overhead, the hidden activation size is set to $R^{C/r\times 1 \times 1}$, where $r$ represents the parameter reduction ratio.
Then, we merge two kinds of feature descriptors $M_{max}$ and $M_{avg}$ via element-wise summation and obtain the channel attention map $M_{c}\in{R^{C \times 1 \times 1}}$ via the Sigmoid activation function~\cite{woo2018cbam}, \emph{i.e.},
\begin{equation}
\begin{aligned}
{M}_{c} &=\sigma(M_{avg} + M_{max})
\end{aligned}
\end{equation}
where $\sigma$ denotes the activation function, which is the logistic function used in practice.

Therefore, we can obtain the attended feature map $F_{c}$ by multiplying the channel attention map $M_{c}$ with the original feature map $F$ as,
\begin{equation}
F_{c} = M_{c} \odot F
\end{equation}
where $\odot$ denotes element-wise multiplication. Here, the channel attention values are broadcasted (copied) along the spatial dimension before the multiply operation.

\emph{Spatial Attention} :
Different from channel attention, spatial attention mainly focuses on mining useful feature regions from the perspective of spatial coordinates. The spatial attention module takes the output of channel attention as input and then returns two-channel descriptions $F_{max}^{c}$ and $F_{ave}^{c} \in {R^{H \times W \times 1}}$ with channel-based global max pooling and global average pooling operations respectively. These descriptions will be fed into a convolutional layer to obtain two kinds of feature descriptors $M_{max}^{c}$ and $M_{avg}^{c} \in {R^{H \times W \times 1}}$. Then, we merge $M_{max}^{c}$ and $M_{avg}^{c}$ together using element-wise summation, and generate the spatial attention map $M_s \in {R^{H \times W \times 1}}$ through a Sigmoid layer as \cite{woo2018cbam},
\begin{equation}
\begin{aligned}
{M}_{s} &=\sigma(M_{avg}^{c} + M_{max}^{c})
\end{aligned}
\end{equation}
Thus, we can obtain the final attended feature map $F_{sc}$ by multiplying the spatial attention map $M_{s}$ with the channel attended feature map $F_{c}$ as,
\begin{equation}
F_{sc} = M_{s} \odot F_{c}
\end{equation}
Similarly, the attention values are broadcasted (copied) along the channel dimension before multiply operation.

\subsubsection{\textbf{Random Selection}}

After obtaining the \emph{attended feature map} and \emph{erased feature map}, how to use them for discriminative feature learning is another key point of the proposed approach. Inspired by the random selection mechanism in previous work \cite{choe2019attention}, we propose to randomly select one from them for subsequent classification in the training process.
On the one hand, the attention module learned in previous iterations will improve the quality of feature map and help the feature dropping module to erase the most discriminative features more accurately. Therefore, we can learn better feature representation than only using the adaptive feature dropping module. On the other hand, our model can capture the local fine-grained features, which can be more effective than attention module in the challenging scenarios where the most discriminative features are inexplicit. The two branches can be trained simultaneously and the features generated from adaptive dropping module and attention module are \emph{opposite}. Concatenating these two features into one representation may weak the attention branch or local feature drop network. Therefore, we select one branch randomly from them to train in each iteration, as suggested in \cite{choe2019attention}. Our experimental results also demonstrate that such alternative learning approach can boost re-ID performance significantly.

\subsection{Training and Testing Phase}

In the training phase, we adopt both label prediction loss and metric learning loss to train the proposed network,
\begin{equation}
\label{finalLoss}
L =  L_{lp} + L_{ml}
\end{equation}
Formally, the label prediction loss $L_{lp}$ is defined as,
\begin{equation}
\label{softmaxLoss}
L_{lp} = \sum\limits_{k=1}^K y_k logS_k
\end{equation}
where $S_k$ is the $k^{th}$ value of the output vector $S$, \emph{i.e.}, the probability of this sample belongs to the $k^{th}$ category. $K$ is the number of categories and $y$ is the ground truth whose dimension is $1 \times K$.

For the metric learning loss, we adopt a soft margin batch-hard triplet loss \cite{hermans2017defense} which aims to increase the distances of negative and anchor samples and decrease the distances of positive and anchor samples. The detailed formulation of this metric learning loss can be written as,
\begin{equation}
\label{smtLoss}
L_{ml} = \sum_{i=1}^P\sum_{q=1}^Nlog(1 + exp(L_{mt}(x_q^i))),
\end{equation}
where
$P$ and $N$ indicate the number of person IDs and images in each ID respectively.
$L_{mt}(x_q^i)$ is defined as \cite{hermans2017defense},
\begin{equation}
\label{mtLoss}
L_{mt}(x_q^i) =\max_{p = 1...N}{D(f(x_q^i), f(x_o^i))} -\min_{\mathclap{\mbox{\tiny$
			\begin{array}{c}
			j=1...P \\ m=1...N \\ j\not=m
			\end{array}$}}}D(f(x_q^i), f(x_m^j))
\end{equation}
Therefore, we have $P \times N$ triplets in a mini-batch. For each triplet sample, we have $x_q^i$, $x_o^i$, $x_m^j$ where $x_q^i$ and $x_o^i$ denote anchor and positive sample respectively and $x_m^j$ denotes the negative sample. $D(\cdot,\cdot)$ is the Euclidean distance function and $f(x)$ represents the feature vector of sample $x$ which is obtained from the last fully connected layer of our network model.

In the testing phase, we jointly utilize features from the global and attention branch as the embedding vector of a given pedestrian image. That is, the local feature drop network is only used in the training phase for robust feature learning.

\subsection{Comparison with Related Works}

The proposed STADB re-ID approach is most related with BDB \cite{dai2019batch}, which proposes a batch dropblock network for person re-ID. Different from BDB \cite{dai2019batch}, the proposed re-ID approach further employs attention scheme \cite{choe2019attention} to adaptively select the attentive regions to erase, which makes our features more discriminative.
STADB is also related with ADL \cite{choe2019attention}. The main differences between STADB and ADL are follows.
First, STADB is designed for person image representation and re-ID tasks, while ADL~\cite{choe2019attention} focuses on weakly supervised object localization.
Second, STADB exploits both channel and spatial attention for feature enhancement while only self-thresholding attention (or spatial) attention is used in ADL~\cite{choe2019attention}.
The channel attention \cite{woo2018cbam} mainly focuses on the different channel information of the input, while the spatial attention \cite{woo2018cbam} mainly focuses on different position information of the input. Therefore, we can attain better feature representation with these two operations.

\section{Experiments}\label{Experiments}

%

\subsection{Datasets and Evaluation Metrics} \label{dataetEvalMetrics}
We evaluate our model on four widely used
{short-term} person re-ID benchmark datasets, including Market-1501~\cite{zheng2015scalable}, DukeMTMC~\cite{zheng2017unlabeled},  CUHK03~\cite{zhong2017re}, and MSMT17~\cite{wei2018person}.
In addition, {to validate the effectiveness of our approach for long-term cloth changing reid scenarios, we also report our results on Celeb-reID~\cite{huang2019celebrities} benchmarks.}
Following the same protocols as previous works~\cite{liu2018pose, zheng2017unlabeled, dai2019batch, Zhong_2018_CVPR}, we evaluate the re-ID performance via two metrics, i.e., mAP and Rank-$k$~\cite{zheng2015scalable}.

\begin{table*}[htp!]
	\centering
	
	\caption{Comparison with other re-ID algorithms on Market-1501 dataset.} \label{resultMarket1501}
	\small
	\setlength{\tabcolsep}{1mm}{
		\begin{tabular}{cccccc}
			\hline \noalign{\smallskip}
			
			Methods & Reference & mAP & Rank-1 & Rank-5 & Rank-10 \\
			\noalign{\smallskip}\hline\noalign{\smallskip}
			SVDNet \cite{sun2017svdnet} & ICCV2017 &	62.1 &	82.3 & 92.3 & 95.2  \\
			HydraPlus \cite{liu2017hydraplus} &	ICCV2017 & 76.9 & 91.3 & 94.5 &- \\
			
			PDC* \cite{su2017pose}& ICCV2017 & 63.451 & 84.14 & 92.73 & 94.92 \\
			Mancs \cite{wang2018mancs}& ECCV2018 & 82.3 & 93.1 &- &- \\	
			HAP2SE \cite{yu2018hard}& ECCV2018 & 69.43 & 84.59 &- &- \\	
			PN-GAN \cite{qian2018pose}& ECCV2018 & 72.58 & 89.43 &- &- \\		
			SGGNN \cite{shen2018person}& ECCV2018 & 82.8 & 92.3 & \color{blue}{96.1} & \color{blue}{97.4} \\
			
			PABR \cite{suh2018Part}& ECCV2018 & 79.6 & 91.7 & -&- \\		
			MGCAM \cite{song2018mask} & CVPR2018 & 74.33 & 83.79 &- &- \\		
			CamStyle+RE \cite{Zhong_2018_CVPR}& CVPR2018 & 71.55 & 89.49 &- &- \\		
			ECN+PSE \cite{saquib2018pose}& CVPR2018 & 80.5 & 90.4 & -&- \\ 		
			HA-CNN \cite{li2018harmonious}& CVPR2018 & 75.7 & 91.2 &- &- \\	
			MLFN \cite{chang2018Multi} & CVPR2018 & 74.3 & 90 & -&- \\		
			DuATM++ \cite{si2018dual}& CVPR2018 & 76.62 & 91.42 & -& -\\		
			DaRe(De)+RE+RR \cite{wang2018resource}& CVPR2018 & 86.7 &90.9 &- & -\\		
			KPM+RSA+HG \cite{shen2018end}& CVPR2018 & 75.3 & 90.1 &- & -\\		
			AOS \cite{huang2018adversarially}& CVPR2018 & 70.43 & 86.49 & -&- \\	
			PCB \cite{sun2018beyond} & ECCV2018 &  {83} & 93.4 &- &- \\	
			BDB\cite{dai2019batch} & ICCV2019 & \color{blue}{86.3} & \color{blue}{94.7}& -&-	\\
			CASN(PCB) \cite{zheng2019re} & CVPR2019 & 82.8 & 94.4 &- & -\\
			BagTricks \cite{luo2019bag}&CVPR2019 & 85.9 & 94.5&-&-\\
			HOReID   \cite{wang2020high} & CVPR2020 & 84.9 & 94.2 &- & -\\
			SNR   \cite{jin2020style} & CVPR2020 & 84.7 & 94.4 &- & -\\	
			
			Top-DB-Net   \cite{quispe2020top} & CVPR2020 & 85.8 & 94.9 &- & -\\

			CtF   \cite{gong2020faster} & ECCV2020 & 84.9 & 93.7&-&- \\
			RDG   \cite{zhuang2020rethinking} & ECCV2020 & 83.6 & 94.3&-&- \\	
			
			\noalign{\smallskip}\hline\noalign{\smallskip}
			STADB (Ours) & &\color{red}{86.7} &\color{red}95.2 &\color{red}97.9&\color{red}98.6\\
			
			\noalign{\smallskip}\hline

		\end{tabular}
	}
\end{table*}

\subsection{Implementation Details} \label{impleDetails}

Due to limited memory of GPU, the down-sampling layers are used in deep networks to reduce the resolution of feature maps (like ResNet50 \cite{he2016deepResidual}). However, for the input pedestrian images, the resolution is low (384 $\times$ 128). If we use all the layers of ResNet50, the response in final features will be weak and many fine-grained features will be lost. To handle this issue, we remove the down-sampling layers after the layer-3 in the backbone network ResNet50 \cite{he2016deepResidual} to make the resolution large enough for recognition. Similar operations can also be found in many other re-id algorithms \cite{luo2019bag, dai2019batch}.
All person images are resized to 384 $\times$ 128.
Our model size is 34.8 M and is trained on a PC with 4 $\times$ GTX-1080 GPUs.
The batch size is 128 with 32 identities in each batch.
We use Adam \cite{kingma2014adam} as the optimizer and the dynamic learning rate is used in the first 50 epochs, \emph{i.e.,} $lr = (0.0001) * (ep / 5 + 1)$, where $ep$ is the index of epoch, $/$ denotes the division operation.
Then, we decay the learning rate to $0.0001$ and $0.00001$ after 200 and 300 epochs respectively.
Our network is trained in 600 epochs. In the local feature drop network, 20\% of the activate values in the feature maps are erased.
In each iteration, the probability of selecting the dropping branch is set as $\rho$ and the probability of selecting the attention branch is $1-\rho$ (we set $\rho$=0.25 in our experiments). In the testing phase, we jointly utilize the feature vectors from both global branch and attention branch as the embedding vector of a pedestrian image.

\begin{table*}[htp!]
	\centering
	\caption{Comparison on the CUHK03 dataset (767/700 split).}\label{resultCUHK03}
	\small
	\setlength{\tabcolsep}{1mm}{
		\begin{tabular}{cccccc}
			\hline \noalign{\smallskip}
			
			& &
			\multicolumn{2}{c}{CUHK03-labeled} &
			\multicolumn{2}{c}{CUHK03-detected} \\
			Methods & Reference & mAP & Rank-1 & mAP & Rank-1 \\
			\noalign{\smallskip}\hline\noalign{\smallskip}
			Mances \cite{wang2018mancs}& ECCV2018& 63.9& 69 & 60.5& 65.5\\
			PN-GAN \cite{qian2018pose}& ECCV2018 & - &79.76& - & 67.65\\
			MGCAM \cite{song2018mask}& CVPR2018& 50.21& 50.14& 46.87& 46.71\\
			HA-CNN \cite{li2018harmonious}& CVPR2018& 41& 44.4& 38.6& 41.7\\
			MLFN \cite{chang2018Multi}& CVPR2018& 49.2& 54.7 & 47.8& 52.8\\
			DaRe(De)+RE+RR \cite{wang2018resource}& CVPR2018& 74.7& {73.8}& {71.6}& 70.6\\
			CASN(PCB) \cite{zheng2019re}& CVPR2019& 68.8& 73.7& 64.4& {71.5}\\
			BDB\cite{dai2019batch} & ICCV2019 & \color{blue}{76.7} & \color{blue}{79.4}& \color{blue}{73.5} &\color{blue}{76.4}	\\
			ISP   \cite{zhu2020identity} & ECCV2020 & 74.1 & 76.6 &71.4& 75.2\\
			Top-DB-Net   \cite{quispe2020top} & CVPR2020 & 75.4 & 79.2 &73.2& 77.3\\
			\noalign{\smallskip}\hline\noalign{\smallskip}
			STADB (Ours) & & \color{red}{80.0}& \color{red}{83.2}& \color{red}{76.2}& \color{red}{79.3} \\
			\noalign{\smallskip}\hline
			
		\end{tabular}
	}
\end{table*}

\subsection{Comparison with State-of-the-art Algorithms} \label{compSOTA}

In this section, we compare the re-ID performance of our method with other state-of-the-art methods on three benchmark datasets, including Market-1501~\cite{zheng2015scalable}, DukeMTMC~\cite{zheng2017unlabeled}, CUHK03~\cite{zhong2017re}, {MSMT17~\cite{wei2018person} and Celeb-reID~\cite{huang2019celebrities} dataset} respectively.

\noindent
\textbf{Results on Market-1501 dataset.}
As shown in Table \ref{resultMarket1501}, BDB~\cite{dai2019batch} achieves 86.3\%, 94.7\% on mAP and Rank-1, respectively; while our method can obtain 86.7\% and 95.2\% on the two metrics respectively. It is also worthy to note that the CASN \cite{zheng2019re} is also developed by combining the local and global features, which achieves 82.8\% and 94.4\% on the mAP and Rank-1. Our method significantly outperforms CASN \cite{zheng2019re}. When obtaining their fine-grained features, CASN \cite{zheng2019re} only focuses on each local area by manually segmenting the feature maps without targeted learning of local features. In contrast, our proposed STADB focuses on learning the non-discriminative features via the local feature dropping network and emphasizing the most discriminative features via the attention module. Therefore, the features extracted by STADB network are more discriminative and thus achieves better re-ID performance. In addition, our model does not require the division of local features, which is more efficient than CASN \cite{zheng2019re}.

\par
\noindent
\textbf{Results on CUHK03 dataset.}	
As shown in Table \ref{resultCUHK03}, the mAP and Rank-1 of our model achieve 80.0\% and 83.2\% respectively on CUHK03-labeled dataset, while 76.2 \% and 79.3 \%  on CUHK03-detected dataset. The baseline approach BDB \cite{dai2019batch} achieves 76.7\% and 79.4\% on the CUHK03-labeled dataset. It is easy to find that our results are 3.3\% and 3.8\% higher than  BDB \cite{dai2019batch} on the CUHK03-labeled dataset. Meanwhile, our results are also  better than BDB \cite{dai2019batch} on the CUHK03-detected dataset, while BDB \cite{dai2019batch} achieves 73.5\% and 76.4\% on the mAP and Rank-1. BDB \cite{dai2019batch} is proposed to randomly erase the feature maps to obtain the local features, however, this simply random dropping operation cannot discard the discriminative regions in original feature maps which may lead to sub-optimal samples for training.


\noindent
\textbf{Results on DukeMTMC dataset. }
As shown in Table \ref{DukeMTMCResults}, our approach achieves 77.1\% on mAP and 89.1\% on Rank-1 on DukeMTMC dataset, which is significantly better than the compared state-of-the-art approaches, including PCB \cite{sun2018beyond} (73.4\% and 84.1\% on the mAP and Rank-1) and BDB~\cite{dai2019batch} (76.0\% and 88.7\% on the mAP and Rank-1). These results consistently promise the performance of our model for fine-grained local feature learning. Note that the proposed approach also outperforms recent HOReID and SNR, as shown in Table \ref{DukeMTMCResults}.

\begin{table}[htp!]
	\centering
	\caption{Comparison on the DukeMTMC dataset.}\label{DukeMTMCResults}
	\small
	\begin{tabular}{cccc}
		\hline \noalign{\smallskip}
		Methods & Reference & mAP & Rank-1 \\
		\noalign{\smallskip}\hline\noalign{\smallskip}
		HAP2SE \cite{yu2018hard} & ECCV2018 & 59.58 & 76.08 \\
		PN-GAN \cite{qian2018pose}& ECCV2018 & 53.2 & 73.58 \\
		SGGNN \cite{shen2018person}& ECCV2018 & 68.2 & 81.1\\
		
		PABR \cite{suh2018Part}& ECCV2018 & 69.3 & 84.4\\
		CamStyle+RE \cite{Zhong_2018_CVPR}& CVPR2018 & 57.61 & 78.32\\
		PSE+ECN \cite{saquib2018pose}& CVPR2018 & 75.5 & 84.5\\
		HA-CNN \cite{li2018harmonious}& CVPR2018 & 63.8 & 80.5\\
		MLFN \cite{chang2018Multi}& CVPR2018 & 62.8 & 81\\
		DuATM++ \cite{si2018dual}& CVPR2018 & 64.58 & 81.82\\
		DaRe(De)+RE+RR \cite{wang2018resource}& CVPR2018 & 80 & 84.4\\
		KPM+RSA+HG \cite{shen2018end}& CVPR2018 & 63.2 & 80.3\\
		AOS \cite{huang2018adversarially}& CVPR2018 & 62.1 & 79.17\\
		PCB \cite{sun2018beyond}&ECCV2018 & 73.4 & 84.1\\
		
		CASN(PCB) \cite{zheng2019re}& CVPR2019 & 73.7 & 87.7\\
		BagTricks \cite{luo2019bag}&CVPR2019 & 76.4 & 86.4\\
		HOReID   \cite{wang2020high} & CVPR2020 & 75.6 & 86.9 \\
		SNR   \cite{jin2020style} & CVPR2020 & 72.9 & 84.4 \\
		
		Top-DB-Net   \cite{quispe2020top} & CVPR2020& 73.5 & 87.5 \\

		CtF   \cite{gong2020faster} & ECCV2020 & 74.8 & 87.6 \\
		RDG   \cite{zhuang2020rethinking} & ECCV2020 & 70.1 & 84.8 \\
		
		BDB \cite{dai2019batch} & ICCV2019 & \color{blue}{76.0} & \color{blue}{88.7}\\
		\noalign{\smallskip}\hline\noalign{\smallskip}
		STADB (Ours) & & \color{red}77.1&\color{red}89.1 \\
		\noalign{\smallskip}\hline
	\end{tabular}
\end{table}

\noindent
{\textbf{Results on MSMT17 dataset. }
To give a more comprehensive experimental analysis, we also report the results on the MSMT17 dataset in Table \ref{MSMT17Results}. Obviously, the baseline method BDB attains 51.5, 78.8, 89.1 on the mAP, Rank-1 and Rank-5, while we achieve 52.3, 79.9, and 89.3 on these three metrics respectively. This fully demonstrates the effectiveness of our proposed adaptive feature dropping module for person re-identification. Compared with other re-ID methods, our results are also comparable even better than them. These results further demonstrate the advantages of our re-ID algorithm. }

\begin{table}[htp!]
	\centering
	\caption{Comparison on the MSMT17 dataset.}\label{MSMT17Results}
	\small
	\begin{tabular}{ccccc}
		\hline \noalign{\smallskip}
		Methods & Reference & mAP & Rank-1& Rank-5 \\
		\noalign{\smallskip}\hline\noalign{\smallskip}
		
		RESNnet50 \cite{he2016deep}& CVPR2016 & 33.9 & 63.2&-\\
		PDC \cite{su2017pose}&ICCV2017 & 29.7 & 58.0& 73.6\\
		GLAD \cite{wei2017glad}& ACM MM17 & 34.0 & 61.4&76.8\\
		
		PCB \cite{sun2018beyond}&ECCV2018 & 40.4 & 68.2&-\\
		
		BagTricks \cite{luo2019bag}&CVPR2019 & 45.1 & 63.4\\
		
		IANet \cite{hou2019interaction} & CVPR2019 & 46.8 & 75.5&85.5 \\
		
		AGW \cite{ye2020deep}  & CVPR2020 & 49.3 & 68.3&- \\
		
		OJMM \cite{zhou2020online}  & CVPR2020 & 43.8 & 74.3&- \\
		
		GASM \cite{he2020guided}  & ECCV2020 & 52.5 & 79.5&- \\
		
		BDB \cite{dai2019batch} & ICCV2019 & 51.5 & {78.8}& {89.1}\\
		\noalign{\smallskip}\hline\noalign{\smallskip}
		STADB (Ours) & & 52.3&79.9&89.3 \\
		\noalign{\smallskip}\hline
	\end{tabular}
\end{table}

\noindent
{
\textbf{Results on Celeb-reID dataset.}
We also test our model on the long-term Celeb-reID dataset \cite{huang2019celebrities} and our method achieves 7.4, 50.2, 65.1 on mAP, Rank-1, and Rank-5. These values are all better than the baseline method. } 

\begin{table}[htp!]
	\centering
	\caption{Comparison on the Celeb-reID dataset.}\label{Celeb-reIDResults}
	\small
	\begin{tabular}{ccccc}
		\hline \noalign{\smallskip}
		Methods  &Reference & mAP & Rank-1& Rank-5 \\
		\noalign{\smallskip}\hline\noalign{\smallskip}

		DenseNet-121 &&5.9 &42.9  & 56.4\\
		ResNet-Mid \cite{yu2017devil}  &ArXiv17& 5.8&43.3&54.6\\
		Two-Stream \cite{zheng2017discriminatively}&TOMM18& 7.8 &36.3& 54.5\\
		MLFN \cite{chang2018Multi} &CVPR18& 6.0&41.4&54.7\\
		
		
		Baseline &&6.8 &48.1  & 63.7\\
		\noalign{\smallskip}\hline
		STADB (Ours) &&7.4& 50.2 & 65.1 \\
		\noalign{\smallskip}\hline
	\end{tabular}
\end{table}

\subsection{Component Analysis}\label{compAnalysis}

In this subsection, we conduct component analysis on DukeMTMC, Market-1501 and CUHK03 datasets to evaluate the effectiveness of each module in our re-ID algorithm. Specifically, six variants of our model are implemented as shown in Table \ref{CA_DukeMTMC}.

$\bullet$ \emph{Global:} global branch used for the feature learning;

$\bullet$ \emph{Drop:} local feature drop network is adopted for robust feature learning;

$\bullet$ \emph{Attention:} attention module is adopted for discriminative feature learning.

As shown in Table \ref{CA_DukeMTMC}, the baseline approach, \emph{i.e.,} only Global branch is adopted for the feature learning, achieves $73.4 \%$, $87.7 \%$ on mAP and Rank-1 on the DukeMTMC dataset. When integrated with Drop branch, the re-ID performance can reach to $76.6\%$, $88.6\%$ on two evaluation metrics.
When we jointly introducing the Global and Attention module, the results reach to $75.4\%$, $88.0\%$. These two experiments validate the effectiveness of the proposed Drop and Attention branch for person re-ID. After integrating these three modules together, the mAP and Rank-1 are further boosted to $77.1\%$, $89.1\%$. Similar conclusions can also be drawn from other re-id benchmarks, which consistently validate the effectiveness of each component in our model.

\begin{table}[htp!]
	\centering
	\caption{Component analysis on DukeMTMC, Market-1501, CUHK03-labeled (CUHK03-LA) and CUHK03-detected (CUHK03-DT) datasets, reported in $mAP | Rank-1$ respectively.}\label{CA_DukeMTMC}
	\scriptsize
	\begin{tabular}{ccc|c|c|c|ccccccc}
		\hline \toprule [0.8 pt]
		\textbf{Global} &\textbf{Drop} &\textbf{Attention} 		&\textbf{DukeMTMC} 	&\textbf{Market-1501} 	&\textbf{CUHK03-LA}		 &\textbf{CUHK03-DT} 		 \\
		\hline
						& 			&\checkmark 				&$74.9|87.5$ 	&$84.4|93.7$	&$72.6|75.4$ 	&$69.5|72.4$	\\
		\hline
						&\checkmark	& 						&$73.0|86.4$	&$84.4|93.5$ 	&$73.2|76.6$	&$70.9|73.5$	\\
		\hline
		\checkmark	& 				& 			 			&$73.4|87.7$ 	&$84.1|94.0$	&$72.9|75.8$ 	&$70.2|73.0$	\\
		\hline
		\checkmark	& 				&\checkmark 		&$75.4|88.0$  	&$84.8|94.4$	&$78.2|80.4$   &$73.9|77.4$	\\
		\hline
		\checkmark	&\checkmark	& 					&$76.6|88.6$  	&$85.7|93.7$	&$78.8|81.6$  	&$74.5|77.8$	\\
		\hline
		\checkmark  &\checkmark &\checkmark 	&$77.1|89.1$	&$86.7|95.2$  	&$80.0|83.2$	&$76.2|79.3$  	\\
		\hline \toprule [0.8 pt]
	\end{tabular}
\end{table}

%
%
%
%
%

\subsection{Ablation Study}

In this section, we will first give an analysis on the attention modules in our model. Then, we report the results with different attention mechanisms. Finally, we report the results with various batch sizes and also analyse the influence of other parameters including $\alpha$ and $\rho$. We also report the running efficiency of our model.

{ \textbf{Influence of Attentions in Our Model.}
To evaluate the contributions of attention modules used in our model, we conduct a component analysis and report corresponding results on CUHK03-labeled and CUHK03-detected dataset in Table \ref{Astam_labeled}. Specifically, the baseline method BDB achieves $76.7|79.4$ on mAP and Rank-1 on CUHK03-labeled dataset, while we get $78.9|81.2$ with self-thresholding attention only. This fully demonstrates that the hard sample generation indeed contributes to the person re-ID task. When we further introduce the spatial-attention or channel-attention, the overall performance can be improved to $79.6|82.6$ and $79.4|82.5$. When all three attention modules are used, we achieve $80.0|83.2$ on this dataset. These experiments fully demonstrate the effectiveness of each attention module for final person re-ID results.
On DukeMTMC and MSMT17, BDB achieves $76.0|88.7$, $51.5|78.8$. Meanwhile, these results can be improved to $76.5|88.9$, $51.8|79.3$ with the self-thresholding attention used. 
 When all the attention modules are used, the results can be further improved. From the aforementioned analysis, we can observe that the proposed self-thresholding attention and the used spatial, channel attention all contribute to our final results.
}

\begin{table}[htp!]
	\centering
	\tiny
	\caption{Ablation study of various attention models on CUHK03-labeled, CUHK03-detected, Market1501, DukeMTMC, MSMT17. CUHK03-labeled and CUHK03-detected are short for CUHK03-LA and CUHK03-DT. self-threshold attention, spatial-attention, channel-attention is short for STA, SA, CA, respectively. \emph{mAP} and \emph{Rank-1} are used as the evaluation metric for these datasets. }\label{Astam_labeled}
	\begin{tabular}{cccc|cccccccccc}
		\hline \toprule [0.8 pt]
		\textbf{BDB}  &\textbf{STA}  &\textbf{SA} &\textbf{CA} &\textbf{CUHK03-LA} &\textbf{CUHK03-DT}  &\textbf{Market1501} &\textbf{DukeMTMC} &\textbf{MSMT17}		 \\
		\hline
		  \checkmark  &  & &  											&$76.7|79.4$ 	&$73.5|76.4$	&$86.3|94.7$	&$76.0|88.7$		&$51.5|78.8$			\\
		  \hline
		  &\checkmark &  &   											&$78.9|81.2$ 	&$75.1|77.9$ 	&$86.2|94.8$	&$76.5|88.9$ 		&$51.8|79.3$ 			\\
		  \hline
	      &\checkmark & \checkmark   &  						&$79.6|82.6$ 	&$75.8|78.8$   &$86.6|95.0$ 	&$76.8|89.0$		&$52.1|79.6$ 			\\
	      \hline
	      &\checkmark &  & \checkmark  						&$79.4|82.5$	&$75.7|78.6$	&$86.5|94.9$	&$76.7|88.9$ 		&$52.0|79.5$ 			\\
	      \hline
     	  &\checkmark  &\checkmark  &\checkmark   		&$80.0|83.2$ 	&$76.2|79.3$	&$86.8|95.2$ 	&$77.1|89.1$ 		&$52.3|79.9$ 			\\
		\hline \toprule [0.8 pt]
	\end{tabular}
\end{table}

{
\textbf{ Results of Different Attention Models.} In this section, we test the overall performance with other attention models to replace the CBAM, including Triplet-attention \cite{misra2020rotate}, SENet  \cite{hu2019squeeze} and SRM  \cite{lee2019srm}. From Table \ref{Dmaa_Market1501}, it is easy to find that the overall performance is relatively stable on all three used evaluation metrics.
}

\begin{table}[htp!]
	\centering
	\caption{Result of different attention models on the Market-1501 dataset}\label{Dmaa_Market1501}
	\small
	\begin{tabular}{cccc}
		\hline \noalign{\smallskip}
		Attention  &mAP&Rank-1 & Rank-5  		 \\
		\noalign{\smallskip}\hline\noalign{\smallskip}
		Triplet-attention  &86.5  & 94.9 &98.0	\\
		SENet              &86.4  &  94.8&98.0	\\
		SRM                &86.7   &  94.7&97.9	\\
		CBAM                &86.7   &  95.2&97.9	\\
	
		\noalign{\smallskip}\hline
	\end{tabular}
\end{table}

{
\textbf{ Influence of Batch Size.} To evaluate the influence of batch size, we report the experimental results with different batch size settings in Table \ref{Dbsa_Market1501}. We can find that our method outperforms the baseline method on all these settings. For example, we improve the baseline from 73.7, 90.1, 96.4 to 79.5, 92.5, 97.2, from 77.4, 91.8, 96.9  to 80.0, 92.2, 96.9, when the batch size are 16 and 32 respectively. These experimental results further  demonstrate the effectiveness of our proposed model for person re-ID.
}

\begin{table}[htp!]
	\centering
	\caption{ Results of different batch sizes on the Market-1501 dataset}\label{Dbsa_Market1501}
	\small
	\begin{tabular}{ccccccc}
		
		\hline \noalign{\smallskip}
		
			&
		\multicolumn{3}{c}{STADB} &

		\multicolumn{3}{c}{Baseline} \\
		
		Batch size &mAP&Rank-1 & Rank-5  &mAP&Rank-1 & Rank-5 		 \\
		\noalign{\smallskip}\hline\noalign{\smallskip}
		16& 79.5 &92.5 &97.2  &73.7	&90.1	&96.4	\\
		32&80.0& 92.2 &96.9	& 77.4&	91.8&96.9\\
		64&84.3&  94.2&97.5&79.7	&93.0	&97.2	\\
		128 &86.7&  95.2&97.9&83.0	&93.4&97.4	\\
		256&83.8&  93.9 &97.6&80.0&	93.5&	97.4	\\	
		\noalign{\smallskip}\hline
	\end{tabular}
\end{table}

\textbf{ Parameter Analysis. }
The proposed STADB has two main parameters, i.e., $\rho$ and $\alpha$. Here, we test the performance of STADB with different parameter settings. For the parameter $\rho$, we set it from 0.25 to 0.90 and report the results in Table \ref{Dps_Market1501}. We can note that the overall performance is stable and changes from 86.2 to 86.7 on mAP, from 94.4 to 95.2 on Rank-1, from 97.7 to 98.0 on Rank-5. 
For the parameter $\alpha$, we set it as 0.5, 0.6, 0.7, 0.8, 0.9, 1.0, the mAP and Rank-1 can be improved when the $\alpha$ changing from 0.5 to 0.8, as shown in Table \ref{Ddra_Market1501}. Therefore, we can get the best results when the value is 0.8.

\begin{table}[htp!]
	\centering
	\caption{ Parameter analysis of $\rho$ on the Market-1501 dataset}\label{Dps_Market1501}
	\small
	\begin{tabular}{cccc}
		\hline \noalign{\smallskip}
		$\rho$ &mAP&Rank-1 & Rank-5  		 \\
		\noalign{\smallskip}\hline\noalign{\smallskip}
		0.25&86.7& 95.2 &97.9	\\	
		0.30&86.7&  94.7&98.0	\\	
		0.40&86.4&  94.8&97.9	\\
		0.50&86.5&  94.9&97.8	\\
		0.60&86.5&  94.8&98.0	\\
		0.75&86.2&  94.4&97.9	\\
		0.80&86.7&  94.9&98.0	\\
		0.90&86.5&  94.7&97.7	\\
		\noalign{\smallskip}\hline
	\end{tabular}
\end{table}

\begin{table}[htp!]
	\centering
	\caption{ Parameter analysis of $\alpha$ on the Market-1501 dataset}\label{Ddra_Market1501}
	\small
	\begin{tabular}{cccc}
		\hline \noalign{\smallskip}
		$\alpha$ &mAP&Rank-1 & Rank-5  		 \\
		\noalign{\smallskip}\hline\noalign{\smallskip}
		1.0&86.2& 94.7 &97.8	\\
		0.9&86.0&  94.3&97.8	\\
		0.8&86.7&  95.2&97.9	\\
		0.7&85.6& 94.6 &97.5	\\
		0.6&85.5&  94.1&97.5	\\
		0.5&85.3&  94.3&97.7	\\
		\noalign{\smallskip}\hline
	\end{tabular}
\end{table}

{
\textbf{ Efficiency Analysis. }
The whole parameters in our model and the baseline BDB network are 35.81M and 32.27M, respectively. In the training phase, the baseline and our model needs 4 and 5 hours respectively on the Market-1501 dataset. The inference of our model can be finished on the Market-1501 dataset in 17.1650 seconds. For each image, it costs about 1.845 microseconds.
}

\subsection{Visualization} \label{Visualization}
\begin{figure*}[!htp]
	\begin{center}		\includegraphics[width=0.8\textwidth]{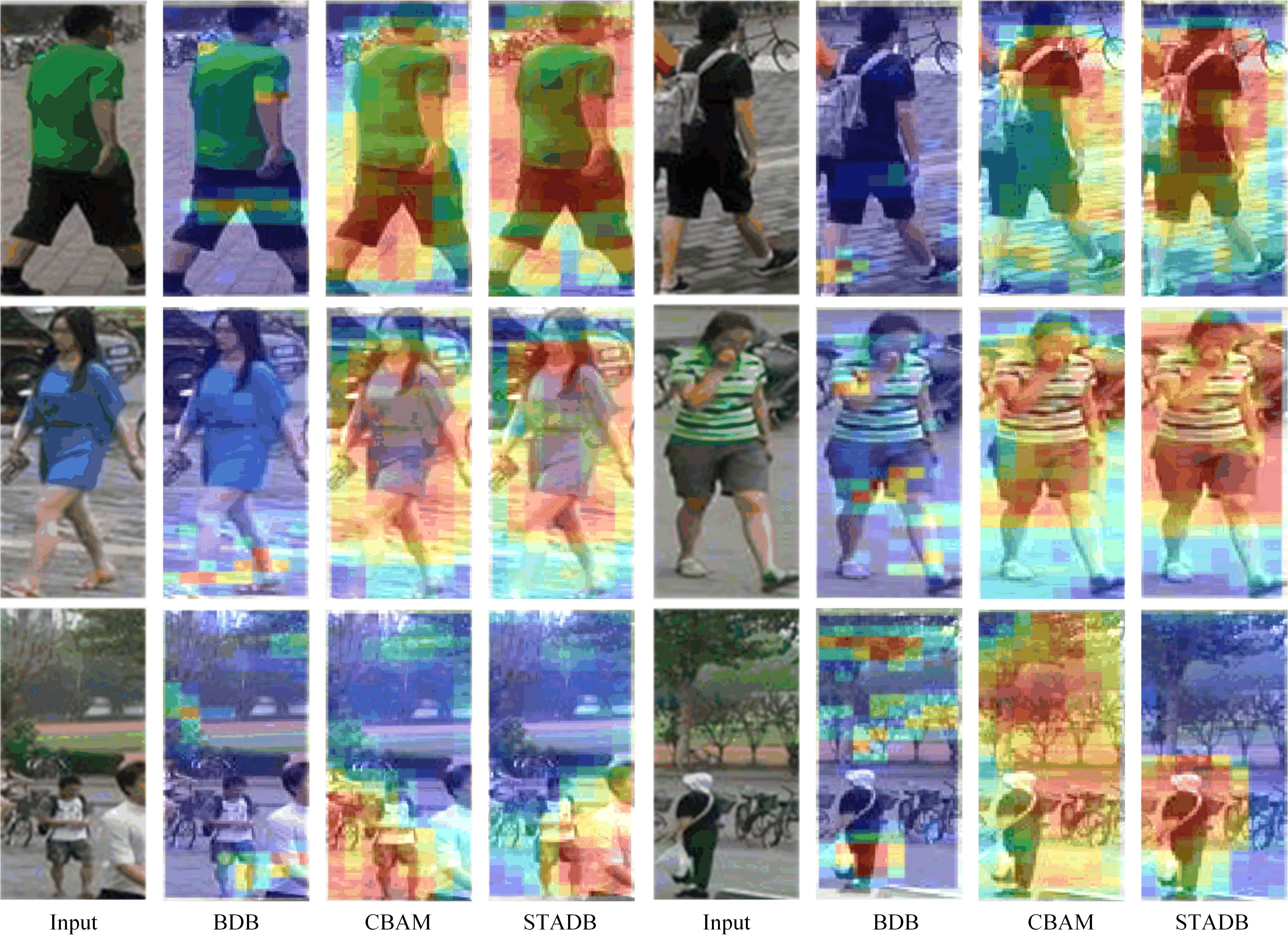}
	\end{center}
	\caption{The visualization of attention maps on the input images of our model and baseline approach and CBAM approach. Red regions represents the area with high response and the Blue color represents the area with few response. }
	\label{fig_attentionVis}
\end{figure*}

\begin{figure*}[!htp]
	\begin{center}
		\includegraphics[width=0.8\textwidth]{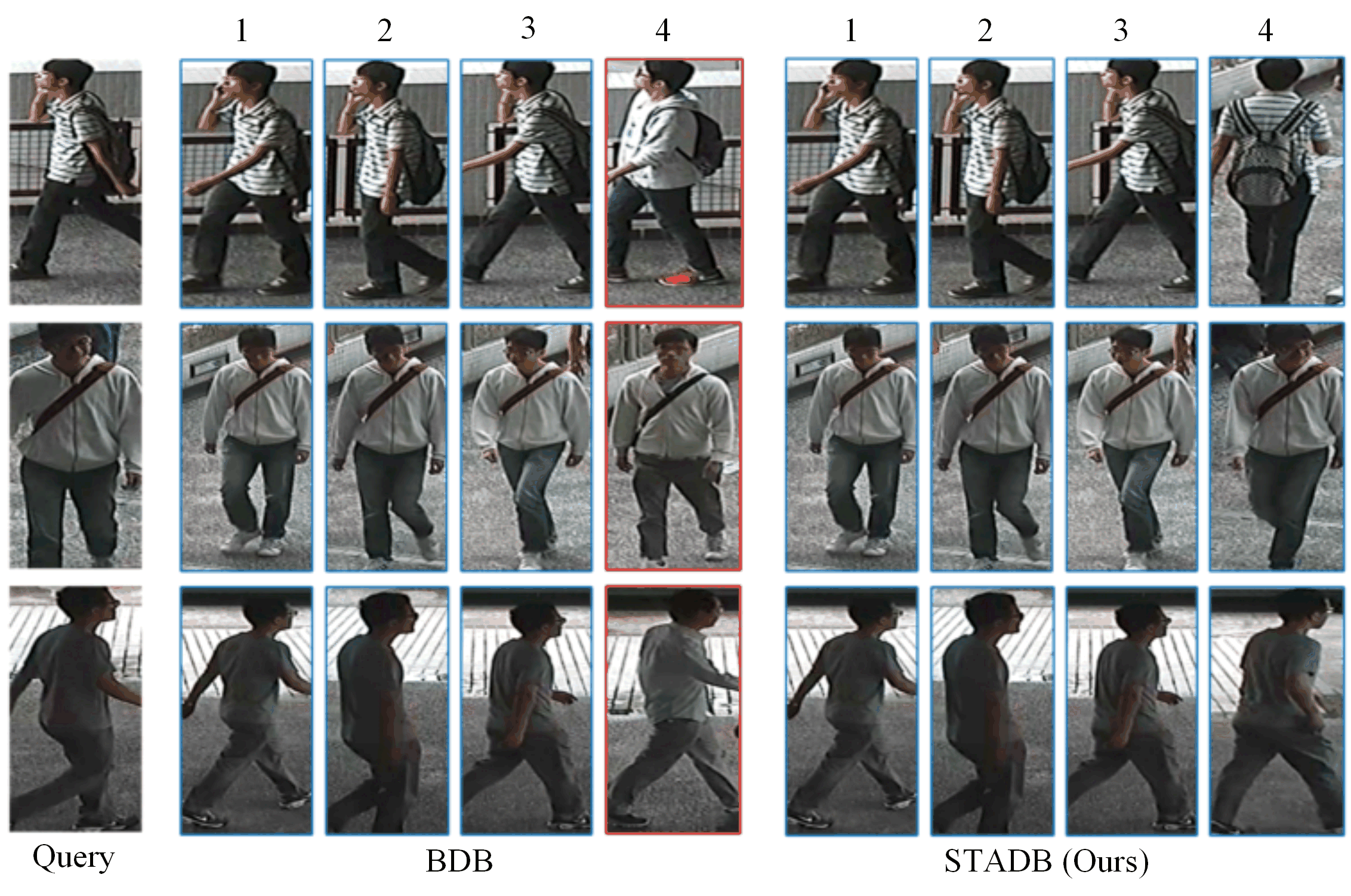}
	\end{center}
	\caption{The visualization of person re-ID results by our proposed algorithm and the baseline method BDB. The images highlighted in Blue are the correct results, the Red denotes false results.}
	\label{fig_resultsVis}
\end{figure*}

In addition to the aforementioned quantitative analysis, we provide some visualizations to show the advantages of our proposed modules. As shown in Figure \ref{fig_attentionVis}, our activation maps all focus on the most discriminative regions of the target person, which responses to the more robust and discriminative feature learning of our method for person re-ID than BDB \cite{dai2019batch}. As shown in Figure \ref{fig_resultsVis}, it is clear that our person re-ID algorithm is significantly better than the baseline method BDB \cite{dai2019batch}.
{We can also find that our STADB is better than the CBAM model \cite{woo2018cbam}.} These visualizations intuitively verify the effectiveness of our proposed fine-grained local feature learning scheme.

\section{Conclusion}\label{Conclusion}
In this paper, Self-Thresholding attention guided Adaptive DropBlock network (STADB) is proposed for robust feature representation learning and person re-ID. Our feature learning framework contains three modules: global branch, local feature drop network and attention branch. To learn more detailed information, we introduce the feature dropping module to erase the most discriminative features. In addition, we utilize the attention mechanism to emphasize the most discriminative local features. The feature dropping module and attention module are trained in an alternative manner via random selection. Extensive experiments on large-scale person re-ID benchmark datasets demonstrate the effectiveness of the proposed STADB re-ID method.

\textbf{Acknowledgement:~~} 
This work is jointly supported by National Nature Science Foundation of China (62076004, 61976002), Postdoctoral Innovative Talent Support Program BX20200174, China Postdoctoral Science Foundation Funded Project 2020M682828.

\small{
\bibliographystyle{elsarticle-num}
\bibliography{reference}
}

\end{document}